\documentclass[10pt,journal,cspaper,compsoc]{IEEEtran}

\usepackage[breaklinks=true,colorlinks,citecolor=blue,linkcolor=blue,urlcolor=blue]{hyperref}
\usepackage{booktabs}
\usepackage{amsmath,amsfonts}
\usepackage{algorithmic}
\usepackage{algorithm}
\usepackage{array}
\usepackage[caption=false,font=normalsize,labelfont=sf,textfont=sf]{subfig}
\usepackage{caption}
\usepackage{textcomp}
\usepackage{stfloats}
\usepackage{url}
\usepackage{verbatim}
\usepackage{graphicx}
\usepackage{multirow}
\usepackage{overpic}
\newcommand{\ourmethod}{WedNet}

\usepackage[nocompress]{cite} 
\usepackage{ragged2e}
\usepackage{cancel}  

\usepackage{geometry}
\usepackage{silence}
\begin{document}

\title{Fast Window-Based Event Denoising with Spatiotemporal Correlation Enhancement}

\author{Huachen Fang, Jinjian Wu,~\IEEEmembership{Member,~IEEE}, Qibin Hou,~\IEEEmembership{Member,~IEEE}, Weisheng Dong,~\IEEEmembership{Member,~IEEE}, Guangming Shi,~\IEEEmembership{Fellow,~IEEE} 

\IEEEcompsocitemizethanks{
\IEEEcompsocthanksitem Huachen Fang, Jinjian Wu, Weisheng Dong, and Guangming Shi are with the
School of Artificial Intelligence, Xidian University, Xi’an, China. (Corresponding
author: Jinjian Wu)
\IEEEcompsocthanksitem Qibin Hou is with VCIP, School
of Computer Science, Nankai University, Tianjin, China.
}}

\markboth{IEEE TRANSACTIONS ON PATTERN ANALYSIS AND MACHINE INTELLIGENCE}%
{Shell \MakeLowercase{\textit{et al.}}: A Sample Article Using IEEEtran.cls for IEEE Journals}



\IEEEtitleabstractindextext{
\begin{abstract} \justifying
Previous deep learning-based event denoising methods mostly suffer from poor interpretability and difficulty in real-time processing due to their complex architecture designs.
In this paper, we propose window-based event denoising, which simultaneously deals with a stack of events while existing element-based denoising focuses on one event each time.
Besides, we give the theoretical analysis based on probability distributions in both temporal and spatial domains to improve interpretability.
In temporal domain, we use timestamp deviations between processing events and central event to judge the temporal correlation and filter out temporal-irrelevant events.
In spatial domain, we choose maximum a posteriori (MAP) to discriminate real-world event and noise, and use the learned convolutional sparse coding to optimize the objective function. 
Based on the theoretical analysis, we build Temporal Window (TW) module and Soft Spatial Feature Embedding (SSFE) module to process temporal and spatial information separately, and construct a novel multi-scale window-based event denoising network, named \ourmethod{}.
The high denoising accuracy and fast running speed of our \ourmethod{} enables us to achieve real-time denoising in complex scenes.
%
%
%
%
Extensive experimental results verify the effectiveness and robustness of our \ourmethod{}. Our algorithm can remove event noise effectively and efficiently and improve the performance of downstream tasks. 
\end{abstract}
}

\maketitle
\IEEEdisplaynontitleabstractindextext
\IEEEpeerreviewmaketitle

\IEEEraisesectionheading{\section{Introduction}\label{sec:introduction}}

%
\IEEEPARstart{E}{vent}-based cameras, including DVS (Dynamic Vision Sensor)~\cite{berner2013240, chen2019live, conradt2009embedded,  lichtsteiner2006128}, ATIS (Asynchronous Time Based Image Sensor)~\cite{2008An} and DAVIS (Dynamic and Active Pixel Vision Sensor)~\cite{berner2013240x180}, are such kind of bio-inspired sensor that capture illuminated change at active pixel unit to trigger a signal. 
In the absence of the computational period of integration and the read-out period (binding all pixels into a frame), it directly detects the log-intensity light variation and creates event only when the alteration exceeds the preset threshold. 
The unique logarithmic differential imaging mechanism brings event camera tremendous benefits of microsecond-level temporal resolution ($\geq800kHz$), high sensing speed ($20us$), and wide dynamic range ($\geq120dB$). 
Due to these alluring characteristics, event camera has gained a lot of academic achievements in many computer vision tasks, such as simultaneous localisation and mapping (SLAM)~\cite{kueng2016low, kim2016real, rebecq2017real, reinbacher2017real, cook2011interacting}, object recognition~\cite{lin2015biologically, amir2017low, lee2014real} and tracking~\cite{vasco2016fast, gehrig2018asynchronous}, optical flow estimation~\cite{rueckauer2016evaluation, gallego2018unifying}, and gesture recognition~\cite{choi2020learning,scheerlinck2020fast,wang2020eventsr}.

\begin{figure}[t]
\centering
\includegraphics[width=\linewidth]{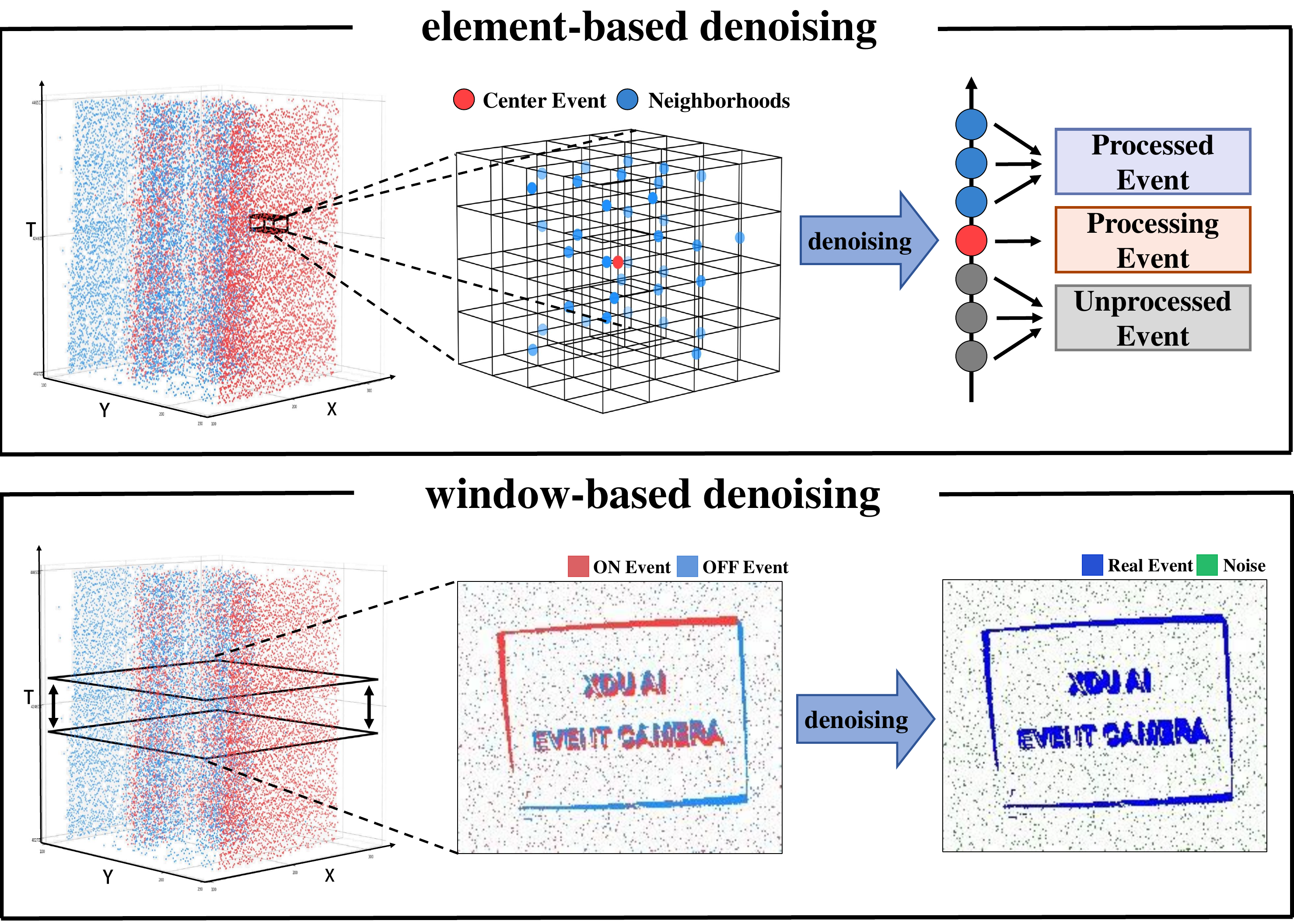}
\caption{\textbf{Top}: Element-based event denoising samples the neighborhoods of the current event and processes the event stream event by event. \textbf{Bottom}: Window-based event denoising method samples a stack of events and labels the event stack at one processing period.}
\label{fig1}
\end{figure}

Though a significant number of academic works prove the potentiality and the advantage of the event-based camera in computer vision tasks, the developing history of the event-based camera is relatively short compared to conventional cameras. Thus, its circuit design is not mature enough.
The hardware deficiency causes fiercer random noise, which leads to the reduction of available communication bandwidth and undoubtedly affects the performance of academic research. 
%
%
%
The most serious noise is background activity (BA). 
BA noise is caused by many hardware factors. 
%
For example, the reset switch fails to close completely, making the leakage currents trigger an unexpected BA noise.
%
%
If a pixel creates an event, it will not produce a BA noise within a short temporal interval. Therefore, event-based camera is relatively unreliable when tracking tiny objects. 
%


\begin{figure}[t]
\centering
\footnotesize
\begin{overpic}[width = 1.0\linewidth]{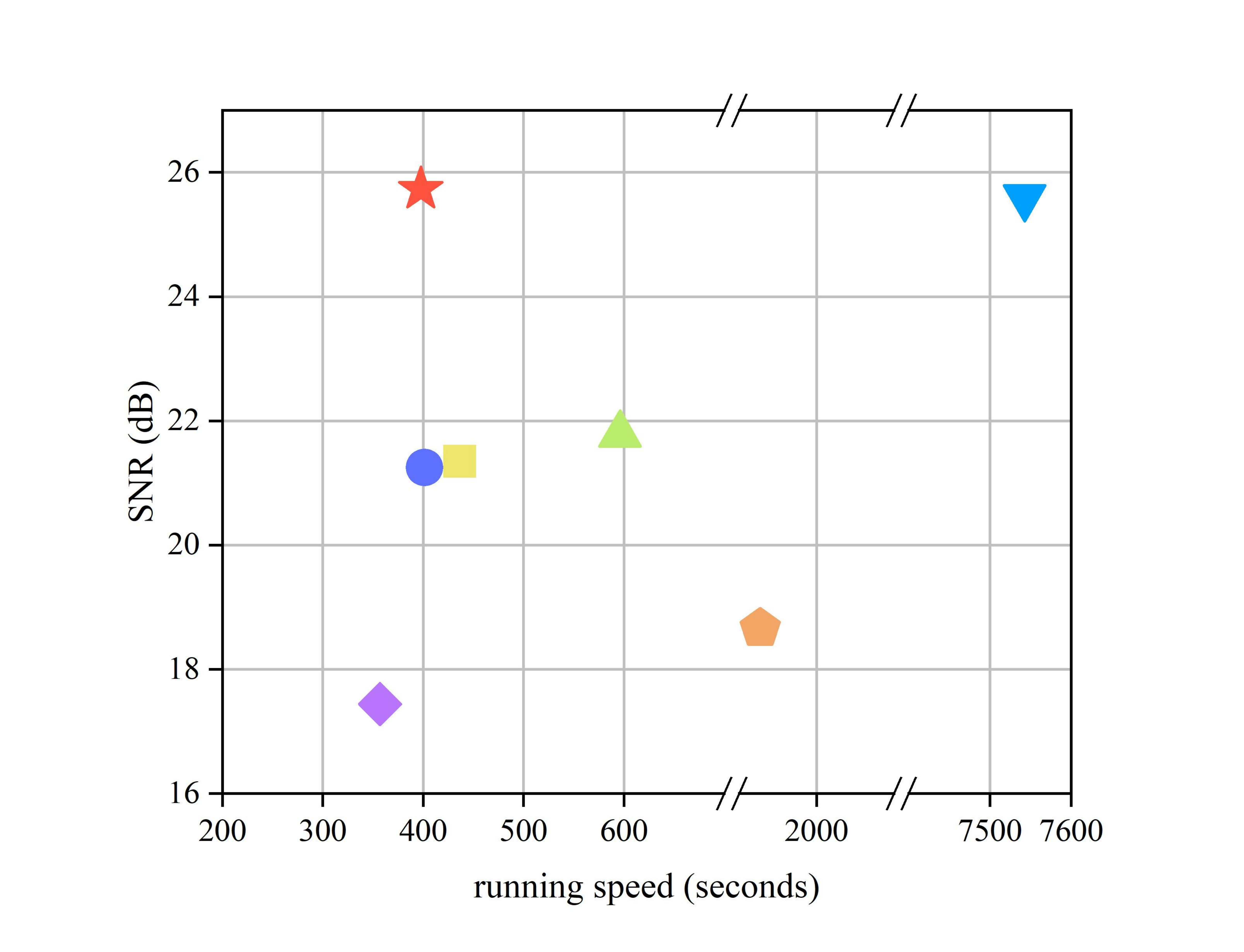}
\put(31,43){BAF~\cite{delbruck2008frame}}
\put(25,34){NNb~\cite{liu2015design}}
\put(28,24){STP~\cite{huang20231000}}
\put(64,25){PUGM~\cite{wu2020probabilistic}}
\put(53,42){EDnCNN~\cite{baldwin2020event}}
\put(66,56){AEDNet~\cite{fang2022aednet}}
\put(28,56){\textbf{WedNet(Ours)}}
\end{overpic}
\caption{Comparisons of SNR score and running time on the DVSCLEAN dataset. The algorithms with higher SNR score and lower running speed have a better denoising performance.}
\label{fig2}
\end{figure}
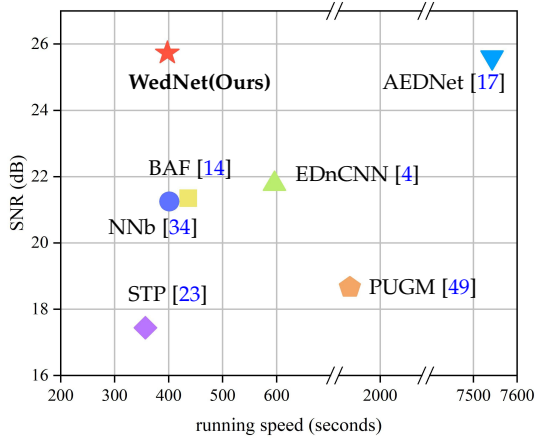

To improve the quality of event-based data, many existing algorithms have attempted to remove the random noise. The main idea of event denoising is to utilize the spatiotemporal correlation. Some design threshold filters to exploit the explicit spatiotemporal correlation, such as BAF~\cite{delbruck2008frame} and NNb~\cite{liu2015design}. They detect the number of events or the temporal difference between two temporal closed events in a spatiotemporal neighborhood. 
These filters show good denoising performance in simple scenes but are subjected to their straightforward judging mechanism, resulting in inferior performance when facing high-noise-ratio scenes.
Later, some researchers design more complicated iterative optimization methods to better utilize  the spatiotemporal correlation among event streams, like inceptive event time surface (IETS)~\cite{baldwin2019inceptive} and guided event filtering (GEF)~\cite{duan2021guided}.
However, these methods concentrate on correlation in mainly one aspect and hence the performance will decrease significantly in some extreme circumstances.
To fully explore the latent spatiotemporal correlation, deep neural networks~\cite{baldwin2020event, duan2021eventzoom, alkendi2022neuromorphic,
afshar2020event, fang2022aednet} are introduced to identify the random noise and get better denoising results.
Nevertheless, existing event denoising networks have the common drawback of low interpretability, making it hard for later researchers to make architectural progress. Also, the expensive computational cost prevents the development of deep neural network in event denoising domain.

In order to solve the problem of the low running speed and low interpretability of previous deep learning based methods, we propose a novel multi-scale window-based event denoising neural network, named \ourmethod{}. 
To be specific, we give a detailed theoretical analysis of how to divide real-world events with noise based on the probability distribution in spatial domain and temporal domain separately.
Due to the unique property of continuation in temporal domain and discreteness in spatial domain, we respectively analyze spatial features and temporal features~\cite{fang2022aednet}. In temporal domain, we use the distribution law to judge the temporal deviation between the central event and other events in the neighbor range. In spatial domain, we select maximum a posteriori (MAP) to define the event denoising optimization problem and utilize the learned convolutional sparse coding to solve the problem. 
Based on our theoretical analysis, we establish the Temporal Window (TW) module and the Soft Spatial Feature Extraction (SSFE) module to extract spatial and temporal features, which offer interpretability and improve the performance of our \ourmethod{}. 
Besides, we use hierarchical set feature learning\cite{qi2017pointnet++} by grouping, sampling, and feature extraction operations to combine the local features with multi-scale receptive fields and achieve window-based event denoising.
Window-based event denoising method as shown in Fig.~\ref{fig1} can handle a stack of temporal-related events simultaneously instead of just one event each time in exsiting element-based denoising, greatly boosting the running speed while keeping good performance.
%
%
Extensive experiments and ablation studies demonstrate the effectiveness and robustness of our method. As shown in Fig~\ref{fig2}, our \ourmethod{} achieves best denoising performace while keeps comparative denoising speed to traditional event denoising methods (STP, NNb and BAF).
To sum up, the main contributions of our paper can be summarized as follows:

\begin{itemize}
\item[•]
We propose a novel multi-scale window-based event denoising network (\ourmethod{}) to speedup the denoising process.
\end{itemize}
\begin{itemize}
\item[•]
We provide a detailed theoretical analysis of separating real-world events from noisy event stream based on probability distribution. 
\end{itemize}
\begin{itemize}
\item[•]
Based on our theoretical analysis, we build the temporal window (TW) module and Soft Spatial Feature Extraction (SSFE) module to separately process temporal and spatial information, which makes our algorithm more interpretable compared to other existing methods.
\end{itemize}


\section{Related Works}
\label{2}

\subsection{Traditional Filter Method}
The main difference between real event and noise is the spatiotemporal correlation with its neighbor events. 
Real events share a high spatiotemporal correlation with their neighbor events while noise is nearly irrelevant to its neighborhoods. 
Liu et al.~\cite{liu2015design} designed the Nearest Neighbor-based (NNb) filter, which checks the number of events in the spatiotemporal neighborhood. 
If the number of events in the neighborhood overcome the predefined threshold, these events will be considered dense enough to pass the filter. Delbruck et al.~\cite{delbruck2008frame} proposed Background Activity Filter (BAF) to filter out noise. This filter checks the timestamp difference between the current event and the most temporal-related event. If the temporal difference exceeds some threshold, the event will be classified as the real one. 
These two algorithms tend to detect BA noise, and they have inferior performance when removing the hot pixel noise. 
Then, the Refractory Period (RP) filter~\cite{czech2016evaluating} was designed to eliminate the impact of hot pixel.
It removes the events with extraordinarily high temporal resolution at the fixed pixel.
Though the above filters are effective in some scenes, the denoising accuracy heavily relies on the choice of the threshold, and these filters need to adjust the threshold manually when the event density varies.
To improve the robustness, Yan et al.~\cite{yan2021adaptive} proposed an adaptive event address map denoising method, which first checks the event density and then adaptively scales the temporal range to adjust the denoising strength. 

The aforementioned methods are offline frameworks. When applying the algorithms to hardware devices, the $O(N^{2})$ memory complexity and the requirement of keeping earlier events challenge hardware deployment.
Khodamoradi et al.~\cite{khodamoradi2018n} proposed a novel online noise filter with $O(N)$ memory complexity, which saves extensive hardware resources. 
Also, Guo et al.~\cite{guo2022low} proposed the fixed and double window filter (FWF$\&$DWF) to save memory and achieve e similar or superior accuracy to the $O(N^{2})$ filter.
However, for both the offline filter or the online designs, there is a common problem that they only care about the conspicuous spatiotemporal correlation but ignore the latent knowledge among the neighborhoods. 
Therefore, the denoising performance is seriously affected when the noise ratio dramatically increases.

\subsection{Iterative Optimization Filter Method}
To better utilize the spatiotemporal correlation, more complicated iterative optimization models are invited. EV-Gait~\cite{wang2019ev} uses the moving consistent plane to filter out inconsistent noise and validate the motion consistency by checking the velocity. Baldwin et al.~\cite{baldwin2019inceptive} assumes that the event edge consists of the original inceptive event (IE) and the following trailing event (TE). IE is considered more informative, so Inceptive Event Time-Surface (IETS) uses iterative local plane to search IE, which only works well on sharp edges. Wu et al.~\cite{wu2020probabilistic} proposed the probabilistic undirected graph model (PUGM) using iterative conditional models (ICM) to minimize the energy function. However, the expensive runtime makes it not applicable for real-time denoising. Duan et al.~\cite{duan2021guided} proposed Guided Event Filtering (GEF), which uses Joint Contrast Maximization (JCM) to associate events with adjacent image frames by a motion model. GEF is based on the linear optical flow assumption. Hence, the performance will be limited when facing the scenarios of non-linear motion and fast illumination variations. These iterative optimization filters comply with merely one criterion, for example, motion consistency and contrast maximization. Many useful spatiotemporal correlation is not included in consideration, so the performance degrades in some particular circumstances, such as highly dim scenes with dramatically increasing noise.

\subsection{Deep Learning-based Method}
Several deep learning-based event denoising methods have also been proposed in recent years. These methods use the feature extraction capability of Deep Neural Networks (DNN) to fully utilize the latent spatiotemporal knowledge. Baldwin et al.~\cite{baldwin2020event} proposed EDnCNN based on 3D convolutional neural network (CNN) to identify the noise with the help of EPM (a kind of label that is calculated by APS and IMU parameters). Duan et al.~\cite{duan2021eventzoom} proposed EventZoom with the backbone of U-Net to incorporate the information of low resolution and high resolution to achieve event denoising and super-resolution. Alkendi et al.~\cite{alkendi2022neuromorphic} proposed a Graph Neural Network (GNN)-driven transformer algorithm to classify every active event pixel in the raw stream into real-log intensity variation or noise. These algorithms show considerable performance compared to traditional filter methods and iterative optimization filter methods. However, deep learning-based methods have two main problems. The first problem is that deep learning-based methods have larger models with more parameters. Hence, there is a substantial computational cost, and it is hard to achieve real-time processes. Secondly, these methods need to be more interpretable since they converge by autonomously learning the difference from the ground truth without theoretical basis and mathematical derivation. They are hard for later researchers to make structural progress.

\begin{figure*}[t!]
\centering
\setlength{\abovecaptionskip}{2pt}
\includegraphics[width = 1.0\linewidth]{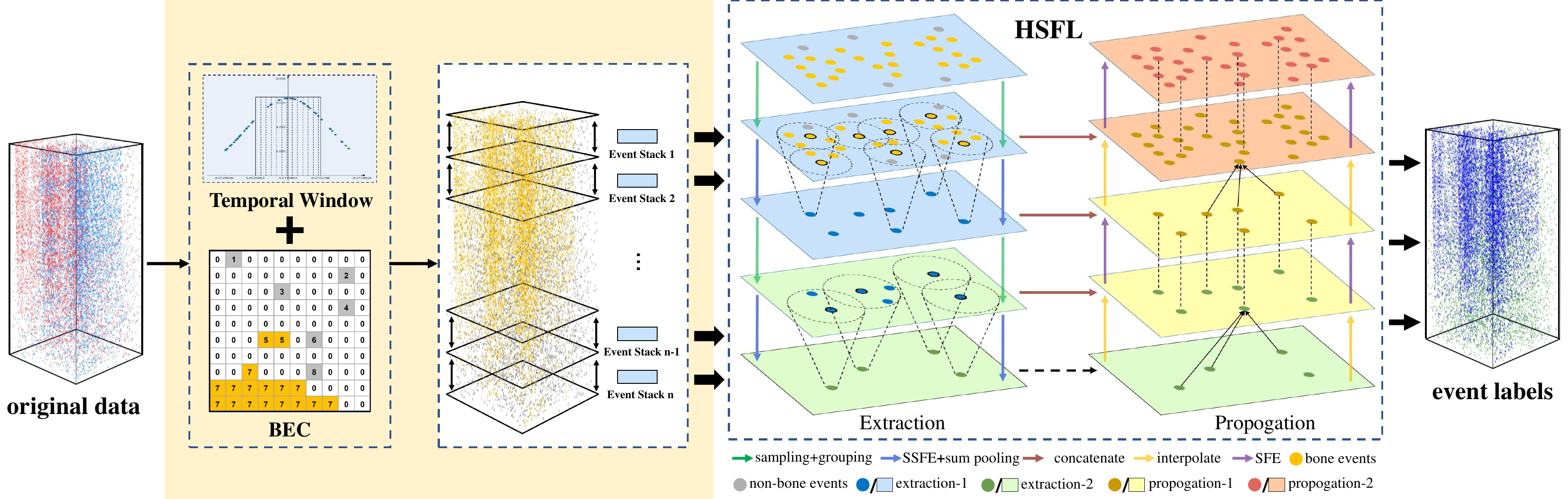}
\caption{Framework of \ourmethod{}. Our \ourmethod{} simultaneously processes a stack of events, significantly improving running speed. We first use the temporal window to divide event stacks and then utilize the BEC module to check the bone events in the event stack. HSFL module is to learn the latent spatial feature consisting of four extraction levels and four propagation levels. Finally, we use fully connected layer to get event labels.}
\label{fig3}
\end{figure*}

\section{Methodology}
\label{3}


The intention of this paper is to solve the problems of low interpretability and low running speed of existing methods.
%
The architecture of our \ourmethod{} can be found in Fig.~\ref{fig3}, which consists of the Temporal Window (TW) module, the Bone Events Check (BEC) module, and the Hierarchical Spatial Feature Learning (HSFL) unit with feature extraction module and feature propagation module. 
We first use the TW module to obtain $w$ temporal-related events. Then, we utilize the BEC module to check the bone events, which helps us prevent feature loss during the sampling operation in the subsequent spatial feature extraction process. 
After acquiring the bone-labeled temporal related events, we put the carried information of these $w$ events in the form of $w*4$ tensor into the Hierarchical Spatial Feature Learning (HSFL) unit. 
The HSFL unit, enlightened by \cite{qi2017pointnet++}, extracts the latent multi-scale spatial knowledge and helps us achieve window-based event denoising.
Unlike the element-based event denoising network that regards event denoising as a point-wise classification task and identifies only one event at each time, our HSFL is a window-based method that can simultaneously process a stack of events, which dramatically increases the efficiency of our algorithm and solves the problem of real-time processing.

\subsection{Theoretical Basis of Event-based Data}
To solve the problem of low interpretability, we give a detailed mathematical derivation of event denoising. We first elucidate the theoretical basis of event-based data. Event-camera simulates the perception mechanism in `what' subpathway of human and non-human primates and abandons the traditional integral imaging mechanism and detects the log-scale bright difference, described as:

\begin{equation}
    \Omega =\mathrm{log}(\frac{aI_{i}+b}{aI_{i-1}+b}), 
\end{equation}
\noindent
where $I_{i}$ and $I_{i-1}$ are the absolute light intensities at coordinate $(x_{i}, y_{i})$ with timestamps $t_{i}$ and $t_{i-1}$, respectively. Parameter $a$ is the gain of the log-scale amplifier and $b$ is the offset to prevent $\log(0)$. The logarithmic amplification signal $\Omega$ is then used to judge whether it is intense enough to qualify an output by the comparator, which can be described as:

\begin{equation}
    E_{i}=\Phi(\Omega, \theta )=\begin{cases} +1, \quad\quad\,\, \text{if}\,\,\Omega \ge \theta 
 \\-1, \quad\quad \text{if}\,\,\Omega \le  -\theta
 \\\,\,0,\quad\quad\quad\,\,\,\,  \text{else} \quad

\end{cases} 
\end{equation}
\noindent
where $\theta$ is the threshold of the comparator. If the absolute value of $\Omega$ overcomes the preset threshold $\theta$, the comparator will generate an event with its polarity based on the gradient direction of bright intensity. 
The comparator will output a positive event when the bright intensity increases and a negative one when the bright intensity decreases. 
Finally, the arbitration circuit outputs a quaternion $e_{i}\left (x_{i}, y_{i},t_{i},p_{i} \right )$, including coordinate positions, timestamp, 
and polarity, after the arbitration mechanism that aims to reduce data volume.

The above assumptions are based on the precondition that the output event stream is noise-free. However, the event stream inevitably mixes with random noise $N$ because of the hardware deficiency, such as threshold mismatch and leakage current shown in Fig.~\ref{fig2}. Therefore, the output signal of the event-based camera can be modified as:

\begin{equation}
    S=E+N=\sum_{i=1}^{m}e_{i} (x_{i}, y_{i}, t_{i}, p_{i})+\sum_{i=1}^{m}n_{i} (x_{i}, y_{i}, t_{i}, p_{i}),  
\end{equation}
where $m$ is the event number of the event stream,  $\left \{ e_{i}\right \} _{i=1}^{m} $ and $\left \{ n_{i}\right \} _{i=1}^{m} $ $\in R^{4\times m} $ refer to real event and noise, respectively. 
If index $i$ refers to real event, $n_{i}$ will be the zero vector. Otherwise, if index $i$ refers to noise, $e_{i}$ will be the zero vector. The noise deteriorates the quality of the event stream and poses a negative effect on subsequent tasks. Our main purpose is to remove the random noises and to recover the pure event stream. We use Maximum A-Posteriori Probability (MAP) to model the event denoising problem as follows:
\begin{equation}
    E=\arg\max_{E} \{P( S|E)\ast P(E)  \},
\label{eq4}
\end{equation}
\noindent
where the former term $P(S|E)$ refers to the posterior probability corresponding to noise, and the latter one $P(E) $ refers to the prior probability corresponding to real event. We transform Eq. (\ref{eq4}) into logarithmic form and use negation operation to solve for the minimum value of the function:
\begin{equation}
    E=\arg\min_{E} \{-\log{P(S|E)}-\log{P(E)}\}.
\label{eq5}
\end{equation}

Eq. (\ref{eq5}) is the objective function of the event denoising task. We can recover the real-world events $E$ from the noisy signal $S$ by optimizing the objective function.
The key points are determining the probability distributions of real-world events and noises, and then choosing the proper optimization method to solve the objective function.
The advantage of our objective function is that it can successfully separate real events and noises, and we can individually analyze the probability distribution law of real events and noises based on their unique properties.

Event-based data is a kind of irregular data that is continuous in the temporal domain and discrete in the spatial domain. Due to the lack of the notion of frame, the spatial information is similar to 2D point cloud consisting of a batch of discrete points across the spatial surface, while the timestamps of events permute continuously along the timeline, giving rise to the high temporal resolution property. Hence, we separately analyze the probability distribution in the spatial domain and temporal domain based on their different properties. We elaborate temporal and spatial denoising processes in Section~\ref{3.2} and Section~\ref{3.3} in detail, respectively.

\subsection{Temporal Window}
\label{3.2}
In the temporal domain, the timestamps of noises randomly permute. The noise is independent and it is produced by hardware deficiencies. Each noise is irrelevant to other noise or real events. We can use Poisson Distribution~\cite{khodamoradi2018n} to describe the temporal information of noise:
\begin{equation}
    P\left \{ N\left ( t \right ) = n \right \} = \frac{(\eta  t)^{n} }{n!} e^{-\eta  t}.
\label{eq6}
\end{equation}
Eq. (\ref{eq6}) gives the probability of an independent pixel generating $n$ noises within the temporal range of $t$, where $\eta $ is the noise rate of the camera. The parameter $t$ does not represent an exact timestamp but indicates a temporal range.
According to Eq. (\ref{eq6}), the possibility of a pixel generating a noise at timestamp $t$ only relates to the noise rate of the camera.

The real events originate from the bright variance caused by object movements. Because of the nature of the high temporal resolution of the event-based camera, the movement of the object will create a large number of real events to depict the instantaneous contour of the moving object. These real events share a high temporal correlation. The tightness of the timestamps reflects the level of temporal correlation. The closer the timestamp of the current event to the center event is, the higher the possibility this current event is a real event. The center event is the temporal average event among the events that depict the current movement. Gaussian Distribution has the property that its probability density function reaches its maximum value at the mean position and exhibits a mirror symmetric attenuation relationship on both sides. Hence, we use the Discrete Gaussian distribution~\cite{canonne2020discrete} to describe the temporal information of real events. 
The probability distribution law can be written as:
\begin{equation}
    \forall t \in \left ( t_{min}, t_{max}   \right ) , p\left \{ t \right \} =\frac{e^{-\frac{(t-t_{\mu})^{2}}{2\sigma ^{2}} } }{ {\textstyle \sum_{t_{k}\in \left ( t_{min}, t_{max}   \right ) }e^{-\frac{(t_{k}-t_{\mu})^{2}   }{2\sigma ^{2} } } } } \\ \\,
\label{eq7}
\end{equation}
where $t_{min}$ and $t_{max}$ is the minimum and maximum timestamp of the event set depicting the current movement, $t_{\mu}$ and $\sigma$ is the temporal mean and the temporal variance of the event batch. In contrast to the probability distribution of noise, the parameter $t$ in Eq. (\ref{eq7}) refers to an exact timestamp. Eq. (\ref{eq7}) assesses the temporal deviation level of the current event among the event batch. The current event that is temporally closer to the center event is more likely to be generated by the real movement. Hence, the event with a higher $P(t)$ is more likely to be judged as a temporal-related event. We apply the normalization operation (divide by the sum of all exponential terms) to introduce the relative temporal relation between the current event and events in the event batch, which can help us better explore the latent temporal information among the event batch. The normalization operation also supports us to achieve the condition that $ {\textstyle \sum_{t\in (t_{min}, t_{max}) } p(t)}=1$. Compared to the probability of noise generation that only relates to the noise rate $\lambda$, the probability distribution of a real event relates to the temporal similarity between the current event and other events in the event batch, which is consistent with the previous hypothesis of the difference between noise and real event.

Note that here we use discrete probability distribution instead of continuous probability distribution to describe the temporal information.
It seems contradictory with the previous analysis that event-based data is consistent in the temporal domain.
The temporal information of event-based data is indeed continuous since the temporal resolution is extremely high. 
However, the camera usually combines with a sampling mechanism during the imaging process, such as the arbitration module.
The sampling mechanism is used to reduce the data volume and increase the speed of imaging.
Although the temporal information is approximately continuous in the camera acquisition process, the event-based output data will be discrete after the arbitration mechanism.
However, the temporal information is still non-homogeneous with spatial information. Therefore, we still need to analyze them separately.

Based on the above analysis, we design the temporal window (TW) module to filter out events with low temporal correlations. Our temporal window module can be described as:
\begin{equation}
\begin{aligned}
     \hat{S}\left ( x,y,t \right )= \{ S\left ( x,y,t \right )\mid \forall t\in \left ( t_{min}, t_{max}  \right ),  \\p(t)\ge p(t_{\mu }-t_{lim}  )  \},
\end{aligned} 
\end{equation}
where $p(t)$ is the probability distribution in Eq. (\ref{eq7}) and $t_{lim}$ is the threshold to judge the temporal correlation. Our TW module retains the events with the timestamps between $(t_{\mu}-t_{lim})$ and $(t_{\mu}+t_{lim})$. These events are considered temporal correlated enough to pass the temporal filter. The denoising intensity is determined by the threshold $t_{lim}$. If $t_{lim}$ is higher, it will reserve more events. Otherwise, more events will be judged as noises and then filtered out. To rationally set $t_{lim}$, we utilize the adaptive threshold in \cite{fang2022aednet}:
\begin{equation}
    t_{lim}=\frac{t_{max}-t_{min} }{\left \lfloor \frac{M}{L}  \right \rfloor },   
\label{eq9}
\end{equation}
where $M$ is the event number of the event batch. Eq. (\ref{eq9}) assumes that averaging $L$ events are sufficient enough to describe a complete transient movement and events generated within $t_{lim}$ are temporally related. The parameter $L$ relates to the hardware configuration of the camera and the complexity of the scene. 

\subsection{Soft Spatial Feature Extraction Module}
\label{3.3}
In the spatial domain, noise is randomly produced among the spatial surface. The spatial information of BA noise is similar to the position information of the Gaussian noise in the conventional image data. Therefore, we use the Gaussian distribution to describe the spatial information of noise:
\begin{equation}
    f(\tilde{N})=\frac{1}{\sigma _{n}\sqrt{2\pi}}e^{-\frac{\tilde{N}^{2} }{2\sigma _{n}^{2} } } ,
\end{equation}
where $\tilde{N}\in R^{m*k} $ refers to the spatial feature of noise $N$.

The spatial information of the real events corresponds to the motion state of a moving object. A dynamic object will leave a locomotive trajectory and event-based camera captures the travelling contour. Hence, we can obtain the geometric shape information by aggregating temporal-related events. If we transform the events into a frame, we can get the edge contour image that describes the traveling trajectory of the moving object. The Generalized Gaussian Distribution (GGD) could be used to analyze the statistical properties of object geometric information. Therefore, we utilize the GGD to describe the spatial information of the real event:
\begin{equation}
    f(\tilde{E})=\frac{\gamma }{2\beta\Gamma(1/\gamma ) } e^{-(\frac{\left | \tilde{E} \right | ^{p}  }{\beta}) } ,
\end{equation}
where $\tilde{E}\in R^{m*k} $ refers to the spatial feature of real event $E$, $\Gamma$ is the gamma function, and $\gamma$ is the shape parameter.

Based on the previous probability density function, we can get the probability distribution function by $\int_{- \infty }^{x} f(t)dt$ and obtain the following relationship:
\begin{equation}
    P(\tilde{N})\sim e^{-\frac{\tilde{N}^{2} }{\sigma_{n}^{2} } },
\label{eq12}
\end{equation}
and
\begin{equation}
    P(\tilde{E})\sim e^{(-\frac{\left | \tilde{E} \right |^{p}  }{\beta} )} .
\label{eq13}
\end{equation}
\noindent
According to the previous analysis, $P(\tilde{N})$ refers to $P(\tilde{S}| \tilde{E})$. We use the real event $E$ and output signal $S$ to describe noise $N$ in Eq.(\ref{eq12}):
\begin{equation}
    P(\tilde{E}|\tilde{S})\sim e^{-\frac{(\tilde{S}-A\tilde{E})^{2} }{\sigma _{n}^{2} } } ,
\label{eq14}
\end{equation}
where $\tilde{S}\in R^{m*k} $ refers to the spatial feature of the output signal $S$, and $A$ refers to the hardware impact that targets the output of real events, such as the refractory period. In the ideal situation, $A$ should be an identity matrix. However, the immature hardware design fails to achieve theoretical replication.

With Eq.(\ref{eq13}) and Eq.(\ref{eq14}), we can update Eq.(\ref{eq5}) as:
\begin{equation}
    \hat{E}=\arg\min_{\tilde{E}} \frac{( \tilde{S}-A\tilde{E})^{2} }{\sigma_{n}^{2} } +\frac{\left | \tilde{E} \right | ^{p} }{\beta } +c ,
\label{eq15}
\end{equation}

\begin{figure}[t!]
\centering
\includegraphics[width = 1.0\linewidth]{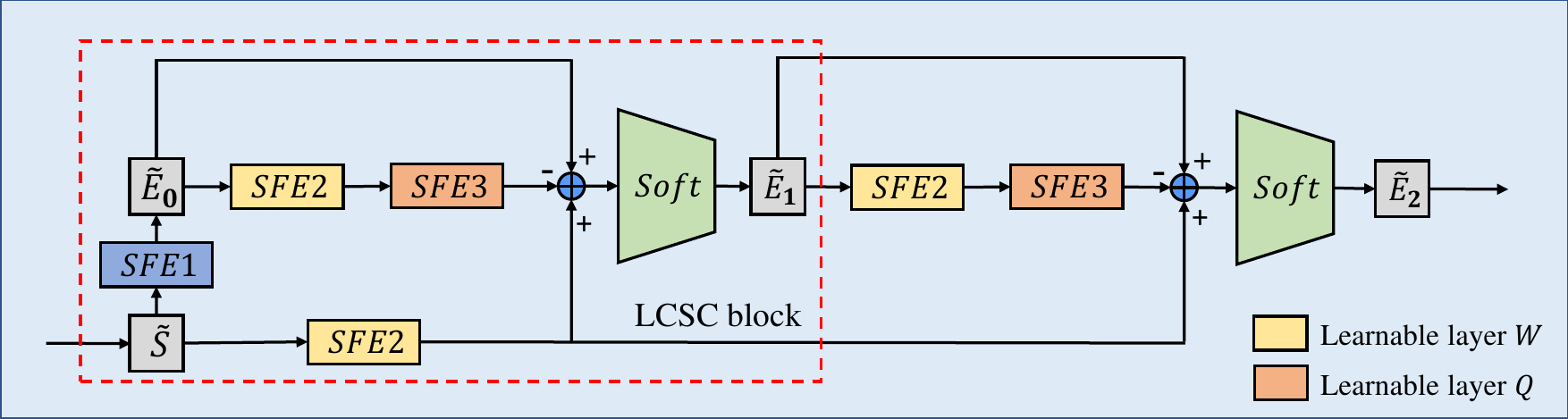}
\caption{Structure of our SSFE module.}
\label{fig4}
\end{figure}
\noindent
where $c$ is the constant term originated from the logarithmic operation. Here, we set $p$ as 1 and organize Eq.(\ref{eq15}) into the norm form:
\begin{equation}
    \hat{e_{i} }=\arg\min_{\tilde{e}_{i} }\left \| \tilde{S} -\sum_{i}a_{i}\tilde{e}_{i}    \right \|_{2}^{2}  + \lambda \sum_{ i} \left \| \tilde{e}_{i} \right \|   _{1}, \ \lambda=\frac{\sigma_{n}^{2}  }{\beta } ,
\label{eq16}
\end{equation}
where $\tilde{e}_{i}$ and $a_{i}\in R^{k} $ refer to the information of a real event and its hardware impact, respectively, and $\lambda$ refers to the iterative number. Eq.(\ref{eq16}) is the standard convolutional sparse coding problem. We use the iterative soft threshold algorithm to solve this problem. Enlightened by the Learned Convolutional Sparse Coding (LCSC) in~\cite{sreter2018learned}, we solve this problem by the following equation:
\begin{equation}
    \tilde{E}_{j+1}=\mathrm{Soft}_{\lambda }(\tilde{E}_{j}- W*Q*\tilde{E}_{j}+W*\tilde{S} ),  
\label{eq17}
\end{equation}
where $\tilde{E} \in R^{m\times k} $ is the stack of spatial feature $\left \{ \tilde{e}^{i}   \right \}_{k=1}^{m} $, $\tilde{E}_{j}$ is the update of $\tilde{E}$ at the $j$-th iteration, $W$ and $Q$ are the learnable convolutional layers. Based on Eq.(\ref{eq17}), we establish the Soft Spatial Feature Extraction (SSFE) module in Fig.~\ref{fig4} to extract the latent spatial feature. We use the spatial feature embedding (the 1D convolution along event direction) in~\cite{fang2022aednet} as our learnable convolutional layers $W$ and $Q$ to well respect the original property of event-based data. To initialize $\tilde{E}$, we use one SFE module to convert the original event stream $\tilde{S}$ to $\tilde{E}_{0}$. The LCSC block in the SSFE module refers to one iteration in Eq.(\ref{eq17}). We can extend the SSFE module to any number of LCSC blocks. SSFE module provides us the interpretability and better ability to extract the spatial feature. 

\subsection{Hierarchical Spatial Feature Learning}
After solving the problem of low interpretability, we then aims at solving the problem of low running speed.
Unlike tasks such as object classification, which cares about the global feature, event denoising focuses on the local feature. Window-based event denoising method introduces the difficulty of abstracting local features among the entire pixel array. 
Therefore, we use the HSFL unit to progressively abstract multi-scale local features along the hierarchy, which helps us better utilize the local spatial correlation and enables us to achieve window-based denoising. 
Our HSFL comprises four feature extraction levels and four feature propagation levels. The spatial receptive region gradually increases, and the sampling events reduce when the set abstraction level climbs. Each level consists of three steps: sampling, grouping, and the SSFE module. 

To be specific, we first sample $T$ typical events to represent the event batch. To fully cover the event batch in the aspect of the spatial domain, we hope the typical events are disperse as much as possible among the spatial surface.
Therefore, we use the farthest event sampling, where the event $e_{i}$ is the most distant event from $\{e_{1}, e_{2}...,e_{i}\}$.
Then, we set the typical events as the centroids of the local features and group $K$ spatial neighborhoods within the radius $r$. 
The event number in the grouping region varies with the event density. 
In this situation, we set the rest events the same as typical event. After the grouping operation, we get the event set of size $ T \times K \times 4 $ and use the SSFE module to abstract the spatial feature. 
We first translate the event set to the relative form by subtracting the carried information of the typical events. The feature extraction block in the SSFE module is the 1D convolution along the $K$ direction to explore the spatial correlation among the local region and maintain the typical event's independence.
Then the spatial correlations are aggregated to the typical events via the sum pooling, and we get the learned spatial feature $\left \{ f^{(j)}(e_{i}) \right \} \in R^{T\times D} $ as the result of feature extraction level-$j$. 
Four extraction levels are used to gradually abstract the local feature. We set the iterative number $\lambda$ in SSFE to 1 to further increase the speed.
$T$ and $K$ in the four levels are set to $ \left [ 2048,512,64,16 \right ] $ and  $ \left [ 64,32,16,8 \right ] $, respectively.

With the final learned spatial feature $\left \{ f^{(4)}(e_{i}) \right \} \in R^{T_{4}\times D_{4}} $, we use feature propagation modules to produce event-wise features for all original events. Our feature propagation module contains the interpolation operation and the spatial feature embedding (SFE) module. In each feature propagation process, the event stack is first propagating by aggregating the features of the three spatial closet events through inverse distance weight:
\begin{equation}
    f'^{(j)} (e_{i})= \frac{ {\textstyle \sum_{h=1}^{3}w_{h}(e_{i}) f^{(j-1)}(e_{i})  } }{ {\textstyle \sum_{h=1}^{3}w_{h}(e_{i})} }, \quad w_{h}(e_{i})=\frac{1}{d(e_{i},e_{h})^{2} } , 
\end{equation}
where $f'^{(j)} (e_{i})$, $f^{(j-1)}(e_{i})$ and $d(e_{i},e_{h})$ are the propagated events of the j-th propagation level, the propagated typical events of the (j-1)-th propagation level, and the distance between events, respectively. We incorporate the previously learned features of typical events with the propagated events to obtain the propagated typical events. Then, we use SFE module to decode the propagated feature. After four propagation processes, we get event-wise features of the $w$ events and obtain the labels by the fully connected layer.

\begin{figure}[htp!]
\centering
\includegraphics[width = 1.0\linewidth]{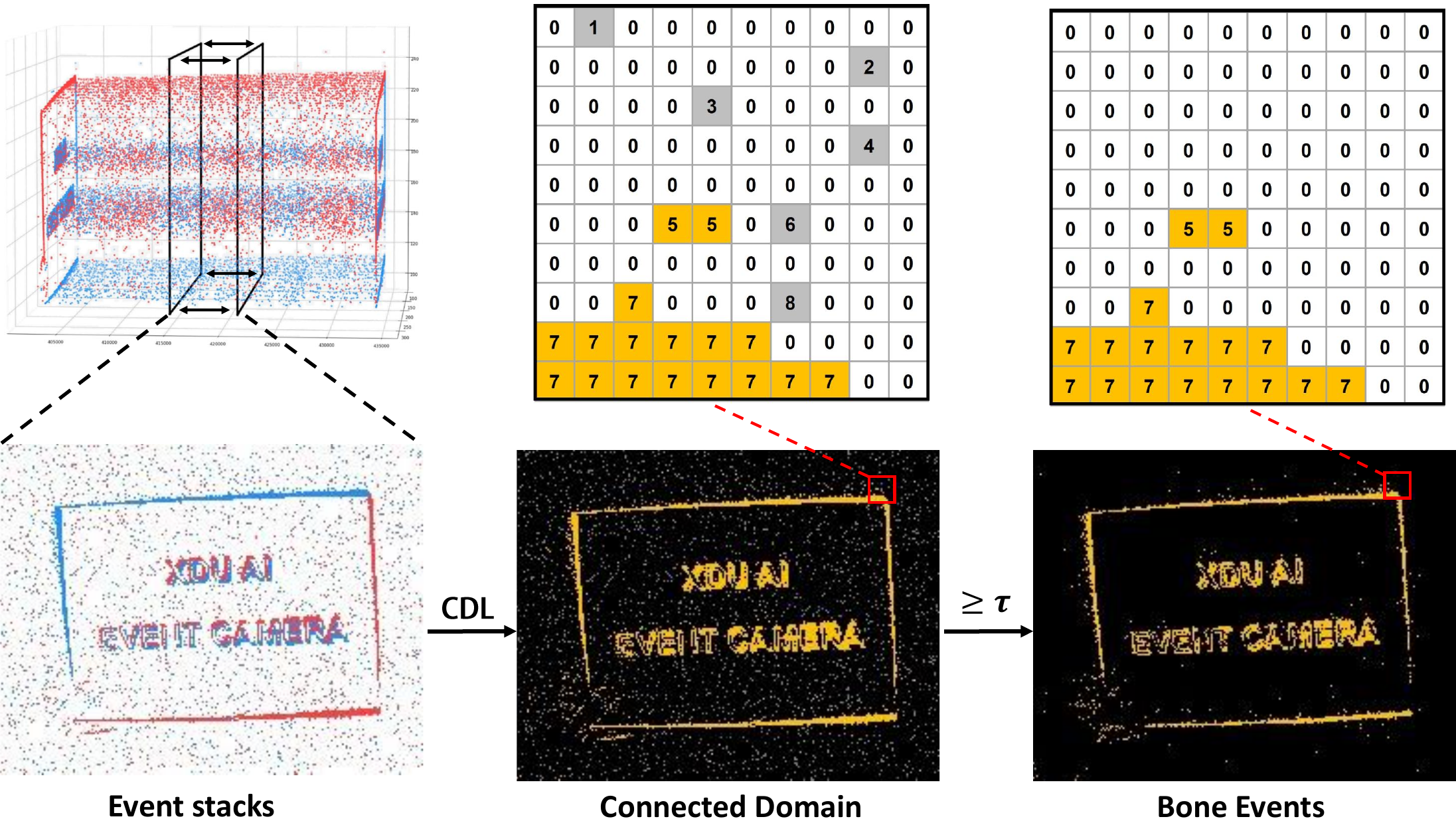}
\caption{Bone Events Check module. A stack of events within $t_{lim}$ is first compressed into a frame. Then, the CDL algorithm (4 neighborhoods) labels the event stack. The event whose connected domain overcomes the threshold $\tau$ is considered a bone event.}
\label{fig5}
\end{figure}

\subsection{Bone Events Check}
During the sampling process in HSFL module, there exists the possibility of sampling the noise as the typical event. The spatial neighborhoods of noise carry little information, which is nearly useless for the local feature extraction around the target object. When the noise ratio grows up, there will be more noise sampling events, and the significant spatial structural knowledge of the moving object may be ignored, which will degrade the denoising performance. To solve this problem, we establish the Bone Events Check (BEC) module to check the bone events as shown in Fig.~\ref{fig5}. We first transform the event batch into a frame and then use the Connected Domain labeling (CDL) algorithm~\cite{shapiro1996connected} to get the connected domain. We judge the connected domain by the predefined threshold $\tau $:
\begin{equation}
    \left \{ \hat{e_{j}}  \right \}_{j=1}^{n'}=\left \{  \left \{ e_{i}  \right \}_{i=1}^{n} \mid C(e_{i}) \ge \tau  \right \} ,
\end{equation}
where $C(e_{i})$ corresponds to the element number in the connected domain containing $e_{i}$ after the CCL algorithm. If the element number overcomes $\tau $, this connected domain can be seen as consisting of the bone events. Otherwise, it fails to get into the sampling process. The threshold $\tau $ should not be too high because we still need the information of noise to differentiate the noise. Therefore, we set $\tau $ as 2 to acquire the best performance. The BEC module can help us better extract the spatial knowledge of the target object while maintaining the spatial information of noise. The effectiveness of Our BEC module is proved in Section~\ref{4.4}.

\section{Experiments}
\label{4}

In this section, we first test our \ourmethod{} to verify the effectiveness and generalization in three public datasets, DVSCLEAN~\cite{fang2022aednet}, DVSNOISE20~\cite{baldwin2020event} and ED-KoGTL~\cite{alkendi2022neuromorphic}. Then 
we compared the running speed of our algorithm with other SOTA methods to prove the competitiveness in real-time process. Finally, we make ablation studies to discuss the validity of our SSFE module and BEC module. For the key parameters, we give the quantitative analysis based on the ablation experiments.

\begin{figure*}[tp!]
\centering
\setlength{\abovecaptionskip}{2pt}
\includegraphics[width = 1.0\linewidth]{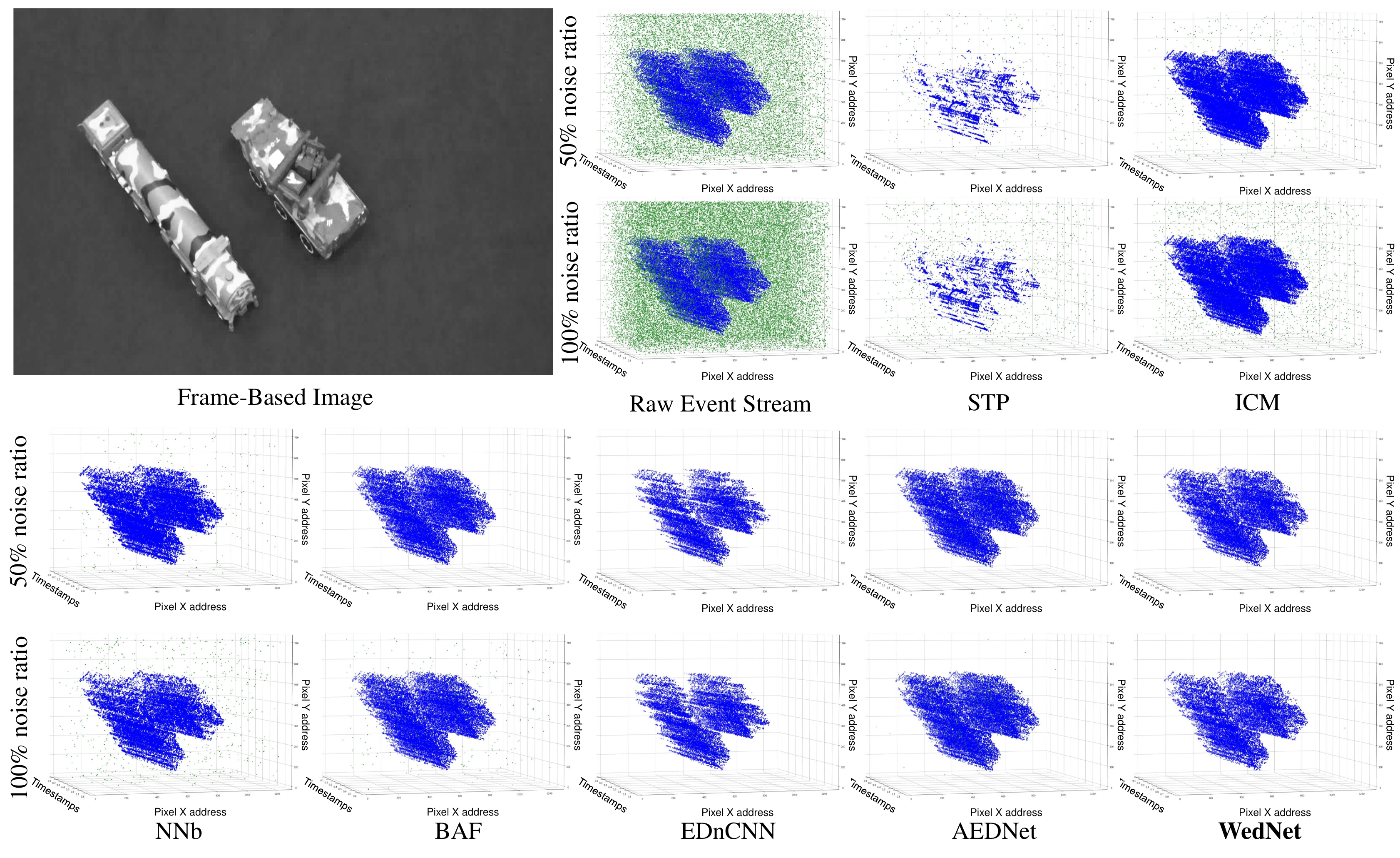}
\caption{Visualization of simulated dataset in DVSCLEAN. Blue points denotes to real event and green points denotes to noise.}
\label{fig6}
\end{figure*}

\begin{figure*}[tp!]
\centering
\includegraphics[width = 1.0\linewidth]{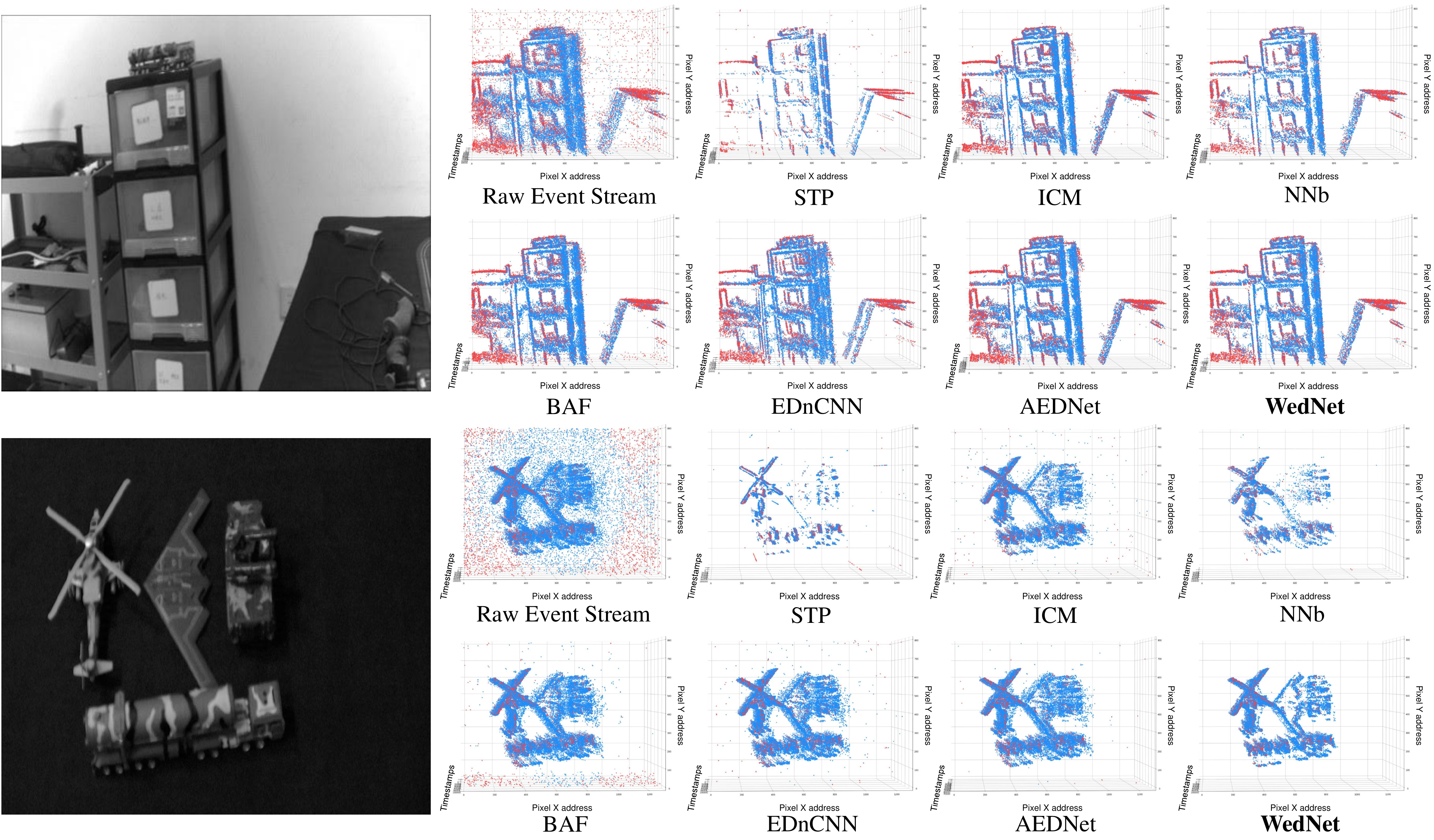}
\caption{Visualization of real-world dataset in DVSCLEAN. Blue point refers to ON event and red point refer to OFF event.}
\label{fig7}
\end{figure*}

\begin{table}[tp!]
\centering
\renewcommand\arraystretch{1.2}
\setlength{\abovecaptionskip}{2pt}
\caption{Comparision of SNR on DVSCLEAN, with the best results in bold and the second best results underlined.}
\label{table1}
\begin{tabular}{cccc}
\hline
                  & \text{50\% noise ratio} & \text{100\% noise ratio} & \text{Average} \\ \hline 
\text{Raw data} & 3                         & 0                          & 1.5              \\ \hline
\text{STP}      & 20.34                     & 14.53                      & 17.44            \\
\text{PUGM}      & 21.64                     & 15.68                      & 18.66            \\
\text{NNb}      & 23.80                     & 18.70                      & 21.25            \\
\text{BAF}      & 23.54                     & 19.16                      & 21.35            \\
\text{EDnCNN}   & 24.75                     & 18.80                      & 21.78            \\
\text{AEDNet}   & \underline{26.11}               & \textbf{25.08}             & \underline{25.60}      \\
\textbf{\ourmethod{}}   & \textbf{26.82}            & \underline{24.65}                & \textbf{25.73}   \\ \hline 
\end{tabular}
\end{table}

\subsection{DVSCLEAN}
DVSCLEAN~\cite{fang2022aednet} is an event denoising dataset consisting of the simulated dataset and the real-world dataset. The real events in the simulated dataset are generated by the ESIM~\cite{gehrig2020video} algorithm, and the noise is artificially added. Hence, the simulated dataset accompanies labels, which can be used to train the model. The simulated dataset has two noise ratio levels, $ 50\% $ and $100\%$ of the number of the simulated-real events. The real-world dataset is the binocular data containing the event stream and frame-based image recorded by the Celex-V camera and the conventional camera. The real-world dataset contains three scene complexity levels: indoor simple scene, indoor complex scene, and outdoor complex scene. There are a total of 49 scenes in the simulated dataset and 44 scenes in the real-world dataset. We use 39 scenes of the simulated dataset as the training set and 10 scenes as the validation set. SNR is used as the denoising metric in DVSCLEAN to benchmark the denoising performance:
\begin{equation}
    \mathrm{SNR}=20\times \log_{10}\frac{M}{N},
\end{equation}
where $M$ and $N$ refer to the number of real events and noise. Higher SNR means better denoising performance.

We compare with other state-of-the-art denoising methods: short-term plasity (STP~\cite{huang20231000}),  probabilistic undirected graph model (PUGM~\cite{wu2020probabilistic}, nearest neighbor (NNb~\cite{liu2015design}), background activity filter (BAF~\cite{delbruck2008frame}), event denoising convolutional neural network (EDnCNN~\cite{baldwin2020event}) and asynchronous
event denoising neural network (AEDNet~\cite{fang2022aednet}). The denoised SNR scores of these seven algorithms can be seen in Table~\ref{table1}.
Our \ourmethod{} achieves the highest SNR score in the $50 \% $ noise ratio scene and the second-best score in the $100 \%$ noise ratio scene. Even if the SNR score of AEDNet in the $100 \%$ noise ratio scene is slightly higher than that of our \ourmethod{}, the average SNR score of our \ourmethod{} is the highest among these seven algorithms. 

The visualization of the denoised event stream can be seen in Fig.~\ref{fig6} and Fig.~\ref{fig7}. We use event stream to visualize the denoising results because the resolution of the Celex-V camera is $1280 \times 800$, greater than $346 \times 260$. If we transform the event stream into a frame, the isolated noise in the denoised event stream is inconspicuous, and it will weaken the visualization effect. Note that there are no labels in the real-world dataset. Therefore, we use polarity to color the events.
We can see PUGM, NNb, and BAF fail to completely remove the isolated noise in high noise ratio scenes. STP and EDnCNN suffer from the problem of removing a lot of real-world events.
The edges of real-world movement become unclear, causing the loss of useful information. 
Besides, STP, PUGM, NNb, BAF, and EDnCNN share a common problem that the denoising performance degrades when the noise ratio increases. 
Our \ourmethod{} not only keeps the real-world structure of events but also removes almost all the isolated noise both in low-noise ratio and high-noise ratio scenes. 
In addition, though AEDNet shows good denoising performance and robustness, the running time is extremely high compared to our \ourmethod{}, which is discussed in Section~\ref{4.5}.

\begin{figure*}[htbp]
\centering
\setlength{\abovecaptionskip}{2pt}
\includegraphics[width = 1.0\linewidth]{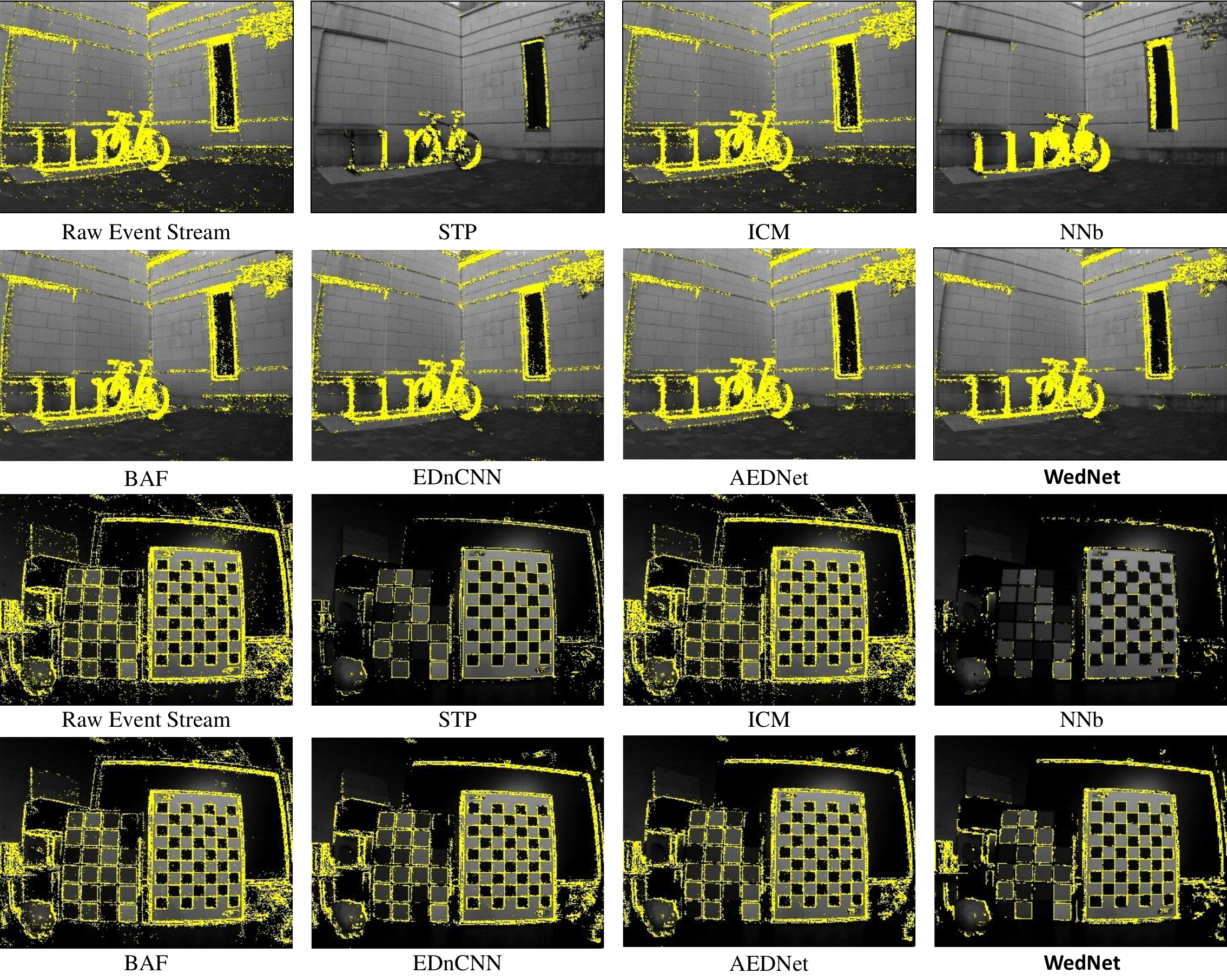}
\captionsetup{justification=centering}
\caption{Visualization of denoising results of SOTA algorithms and our \ourmethod{} on published dataset DVSNOISE20.}
\label{fig8}
\end{figure*}

\begin{figure*}[htbp]
\centering
\includegraphics[width = 1.0\linewidth]{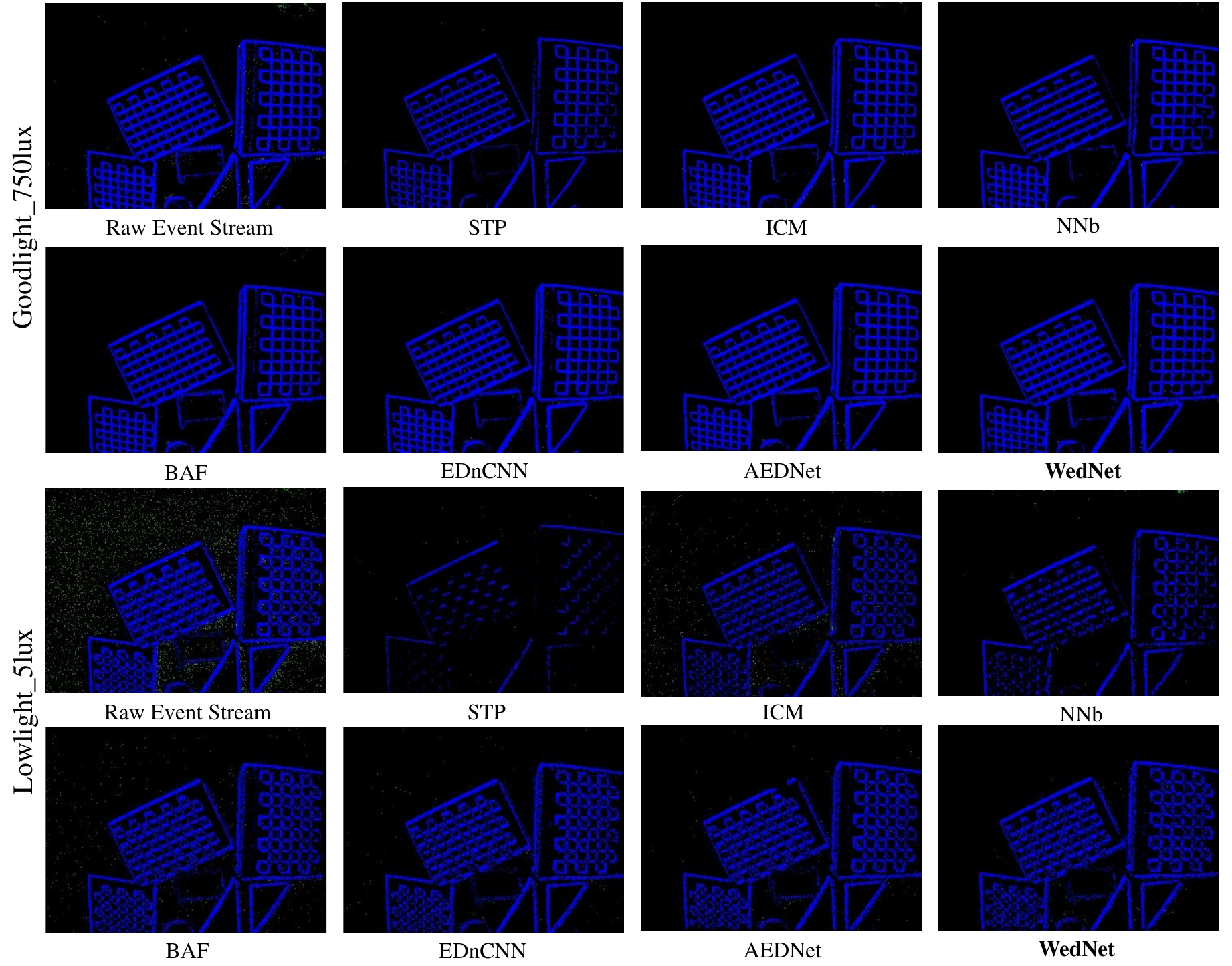}
\caption{Visualization of denoising results of SOTA algorithms and our \ourmethod{} on published dataset ED-KoGTL.}
\label{fig9}
\end{figure*}

\begin{table*}[]
\centering
\renewcommand\arraystretch{1.3}
\setlength{\abovecaptionskip}{2pt}
\setlength\tabcolsep{1.7pt}
\footnotesize
\caption{\centering Comparision of RPMD on DVSNOISE20 dataset, with the best results in bold and the second best results underlined. Smaller RPMD values indicate better denoising performance.}
\label{table2}
\begin{tabular}{c|cccccccccccccccc|c}
\hline
                & \text{Alley} & \text{Bench} & \text{Bigchk} & \multicolumn{1}{l}{\text{Bike}} & \multicolumn{1}{l}{\text{Bricks}} & \multicolumn{1}{l}{\text{ChkFast}} & \multicolumn{1}{l}{\text{ChkSlow}} & \multicolumn{1}{l}{\text{Class}} & \multicolumn{1}{l}{\text{Conf.}} & \multicolumn{1}{l}{\text{LabFast}} & \multicolumn{1}{l}{\text{LabSlow}} & \multicolumn{1}{l}{\text{Pavers}} & \multicolumn{1}{l}{\text{Soccer}} & \multicolumn{1}{l}{\text{Stairs}} & \multicolumn{1}{l}{\text{Toys}} & \multicolumn{1}{l|}{\text{Wall}} & \multicolumn{1}{l}{\text{Avg.}} \\ \hline
\text{STP}    & 169.25         & 136.92           & 157.39              & 194.78                            & 146.82                              & 135.79                                   & 164.79                                   & 238.57                             & 235.12                                  & 98.43                                & 83.62                                & 203.45                              & 120.46                              & 198.46                              & 349.24                            & 203.45                             & 163.97                                \\
\text{PUGM}    & 121.92         & 150.41           & 247.69              & 130.73                            & 150.75                              & 139.92                                   & 148.85                                   & 66.82                              & 214.39                                  & 220.89                               & 123.68                               & 146.50                              & 67.74                               & 85.78                               & 268.78                            & 125.87                             & 149.24                               \\
\text{NNb}    & 186.74         & 42.43            & 106.72              & 65.81                             & 9.67                                & 79.18                                    & 53.25                                    & 138.27                             & 148.36                                  & 93.74                                & 63.84                                & 120.96                              & 30.42                               & 67.98                               & 155.29                            & 205.31                             & 97.99                                \\
\text{BAF}    & 197.52         & 32.48            & 103.42              & 69.73                             & 12.81                               & 73.47                                    & 61.58                                    & 126.35                             & 145.83                                  & 85.75                                & 43.68                                & 137.31                              & 17.59                               & 74.58                               & 161.49                            & 178.51                             & 95.13                                \\
\text{EDnCNN} & 43.29          & 40.78            & 43.52               & 7.34                              & 15.29                               & 25.93                                    & 33.64                                    & 26.41                              & 28.59                                   & 45.17                                & 37.82                                & 46.61                               & 22.75                               & 39.48                               & 45.94                             & 64.71                              & 35.45                                \\
\text{AEDNet} & 26.57          & 18.63            & 65.39               & 6.08                              & 35.84                               & 51.87                                    & 52.26                                    & 11.57                              & 18.65                                   & 19.68                                & 24.17                                & 17.62                               & 13.57                               & 36.18                               & 39.64                             & 45.68                              & \underline{30.21}                          \\
\textbf{\ourmethod{}} & 35.17          & 25.85            & 49.61               & 5.57                              & 24.23                               & 38.85                                    & 33.90                                    & 15.34                              & 14.76                                   & 14.42                                & 19.02                                & 25.84                               & 24.21                               & 40.53                               & 34.11                             & 48.54                              & \textbf{28.12}                       \\ \hline
\end{tabular}
\end{table*}

\subsection{DVSNOISE20}
DVSNOISE20~\cite{baldwin2020event} is collected by the DAVIS346 camera, with a resolution of $346 \times 260$, active pixel sensors (APS), and an inertial measurement unit (IMU). It contains 16 indoor and outdoor scenes. Each scene is captured three times, and therefore, there is totally $48$ sequences with a wide range of motions.
With the help of APS and IMU, we acquire event probability mask (EPM) to represent, which quantifies the plausibility of observing an event within the time window. Because it needs the information of APS, EPM only exists within the exposure time. Relative Plausibility Measure of Denoising (RPMD) is the measuring metric.
Lower RPMD values means better denoising performance.

We select the $2\%$ middle exposure temporal window and test the denoising performances of the alogorithms. The denoising performance can be seen in Fig.~\ref{fig8} and Table~\ref{table2}. Our \ourmethod{} achieves the best RPMD values. The first scene in Fig.~\ref{fig8} has low noise ratio and obscure edges. It tests the ability to maintain the real-world events when removing noises. The second scene has a high noise ratio and scene complexity. It tests the ability to remove the noise near the edge of the moving object. STP has a fiercer denoising ability, therefore, the denoising performance is better in the second scene, while a lot of real-world edges are mistakenly removed in the first scene. PUGM, BAF, EDnCNN, and AEDNet suffer from the high texture contents. They show competent denoising ability in the first scene while having inferior performance in the second scene. Our \ourmethod{} has good denoising visualization effect in both scenes, proving the robustness.

\subsection{ED-KoGTL}
ED-KoGTL is recorded by a DAVIS346C and a Universal Robot UR10 6DOF arm. The neuromorphic camera is mounted on the arm in a front forward position and repeatedly moved along a certain (identical) trajectory under four illumination conditions, particularly \textasciitilde$750lux$, \textasciitilde$350lux$, \textasciitilde$5lux$, and \textasciitilde$0.15lux$. The ground-truth label is obtained by the Known-object Ground-Truth Labeling (KoGTL), which uses the Canny algorithm to extract edge information from APS and labels the detected edge events as real-world events. We test the algorithms on two public illumination conditions, very good light condition (\textasciitilde750lux) and low light condition (\textasciitilde5lux), and use the SNR metric to evaluate the denoising performance. The denoising results can be seen in Fig.~\ref{fig9} and Table~\ref{table3}. Lowlight\_5lux scene 

\begin{table}[]
\setlength{\abovecaptionskip}{2pt}
\renewcommand\arraystretch{1.2}
\centering
\caption{\centering Comparision of SNR on ED-KoGTL dataset, with the best results are in bold and the second best results are underlined.}
\label{table3}
\begin{tabular}{c|cc|c}
\hline
                  & \text{Goodlight\_750lux} & \text{Lowlight\_5lux} & \multicolumn{1}{l}{\text{Average}} \\ \hline
\text{Raw data} & 19.17                      & 10.09                            & 14.63                                \\ \hline
\text{STP}      & 27.14                      & 13.81                            & 20.48                                \\
\text{PUGM}      & 25.11                      & 16.74                            & 20.93                                \\
\text{NNb}      & 26.28                      & 16.10                            & 21.19                                \\
\text{BAF}      & 26.30                      & 17.39                            & 21.84                                \\
\text{EDnCNN}   & 27.16                      & 20.35                            & 23.76                                \\
\text{AEDNet}   & \underline{29.56}                & \underline{21.51}                      & \underline{25.54}                          \\
\textbf{\ourmethod{}}   & \textbf{30.25}             & \textbf{23.69}                   & \textbf{26.97}                       \\ \hline
\end{tabular}
\end{table}

\noindent
is accompanied with more noises compared to the Goodlight\_750lux scene because the event-based camera tends to produce more noises under dim light. It is observed that our \ourmethod{} outperforms other SOTA algorithms. It improves the SNR metric by $11.08dB$ and $13.6dB$ on the goodlight scene and lowlight scene, respectively.

\begin{figure*}[htbp]
\centering
\includegraphics[width = 1.0\linewidth]{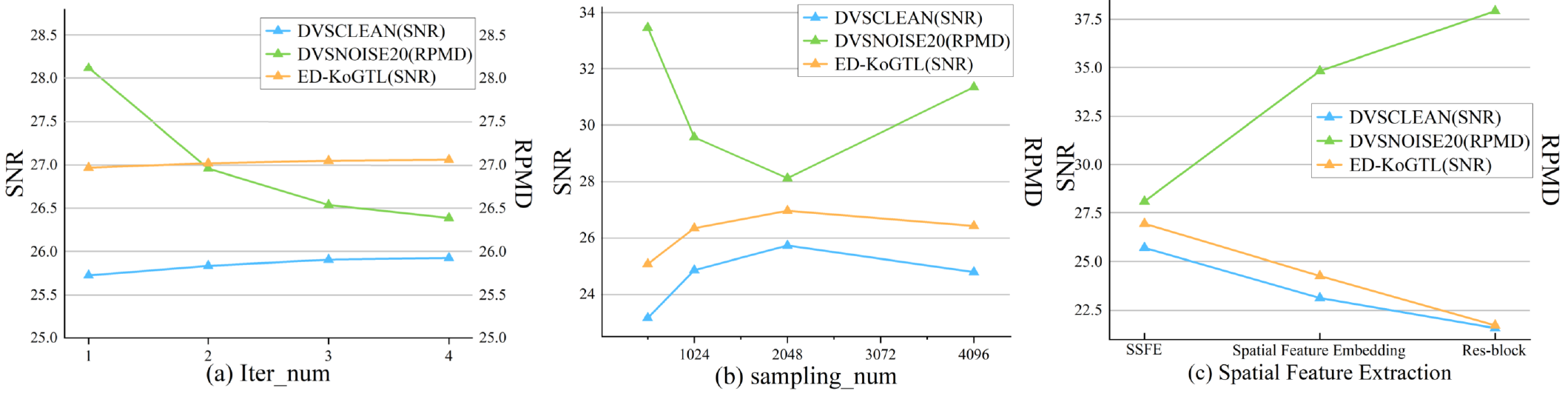}
\caption{Results of ablation study. Notice that lower RPMD values refers to higher denoising performance. (a) Relationship between iterative number in the SSFE module and denoising performance. (b) Relationship between sampling number and denoising performance. (c) The comparison of denoising performance of different spatial feature extraction module.}
\label{fig10}
\end{figure*}

\begin{table}[]
\setlength\tabcolsep{1.7pt}
\setlength{\abovecaptionskip}{2pt}
\renewcommand\arraystretch{1.2}
\centering
\caption{\centering Denoising runtime comparison, with the best results are in bold and the second best results are underlined. (unit: second)}
\label{table4}
\begin{tabular}{c|c|ccc}
\hline
Type                            & Methods         & \multicolumn{1}{l}{DVSCLEAN} & \multicolumn{1}{l}{DVSNOISE20} & \multicolumn{1}{l}{ED-KoGTL} \\ \hline
\multirow{4}{*}{filter-based}   & STP             & \textbf{356.94}              & \underline{441.76}                   & \textbf{369.25}              \\
                                & PUGM            & 1927.36                      & 1943.66                        & 1871.86                      \\
                                & NNb             & 401.38                       & 485.36                         & 433.74                       \\
                                & BAF             & 436.38                       & 503.51                         & 459.82                       \\ \hline
\multirow{3}{*}{learning-based} & EDnCNN          & 596.12                       & 627.51                         & 604.59                       \\
                                & AEDNet          & 7542.55                      & 8379.58                        & 8267.39                      \\
                                & \textbf{WedNet} & \underline{397.68}                 & \textbf{427.54}                & \underline{386.75}                  \\ \hline
\end{tabular}

\end{table}

\subsection{Running speed}
\label{4.4}
Our goal is to solve the problem of the low running speed of the deep-learning-based event denoising method and improve the denoising efficiency. Therefore, we make running time comparison experiments on the three datasets as shown in Table~\ref{table4}. EDnCNN, PUGM and AEDNet spend more running time compared to other algorithms on all three datasets. Our \ourmethod{} takes at least 20 $\times$ less time than other deep learning methods. It overcomes most of the conventional filters (NNb and BAF) and even achieves the fastest on DVSNOISE20, which proves the effectiveness of improving efficiency. The time is recorded by implementing the experiments on a PC with a GEFORCE GTX 3090 Ti GPU.

\subsection{Ablation Study}
\label{4.5}

\textbf{BEC Unit.} In the first ablation study, we explore the importance of our BEC unit on denoising performance. To do this, we make comparison experiments on three datasets. The sampling operation without a BEC unit directly utilizes the farthest sampling. Table~\ref{table6} shows the results with and without the BEC unit. As we can see, the BEC unit indeed helps increase the denoising accuracy, especially in the DVSCLEAN dataset. Because the spatial resolution is higher and there are more non-object areas. Hence, the farthest sampling strategy will include more isolated noise as typical events, which will affect the denoising performance.

\textbf{Iterative number.} Then, we analyze the impact of iterative number $\lambda$ in the SSFE module. As shown in Fig.~\ref{fig10}(a), the denoising performance increases as the iterative number increases (SNR value increases in DVSCLEAN and ED-KoGTL datasets and RPMD value decreases in DVSNOISE20 dataset). However, the increment is negligible, and the increase of the iterative number will inevitably increase the running time. To reduce the parameters of our method and to maintain the denoising efficiency, we set the iterative number as 1.

\textbf{Sampling event number.} 
We also analyze the relationship between the denoising performance and the sampling event number in the first sampling operation. As we can see in Fig.~\ref{fig10}(b), the denoising performance improves as the increase of sampling event number when the sampling number is relatively few since more events include more information. However, when the sampling number is sufficient, the denoising accuracy will not increase and even decrease. This is because too many events have no benefits on local feature extraction. In this paper, we set the sampling number as 2048. 

\textbf{Spatial feature extraction.} We further analyze the effectiveness of our SSFE module. Our SSFE module is established to better extract the spatial feature. We compare with other feature extraction modules, Spatial Feature Embedding module~\cite{fang2022aednet} and Res-block~\cite{he2016deep} on three datasets. The comparison results can be seen in Fig.~\ref{fig10}(c). Even if the spatial feature embedding module has better denoising performance than Res-block, our SSFE module outperforms other feature extraction methods on event denoising tasks, which proves the rationality of our mathematical derivation.

\begin{table}[]
\renewcommand\arraystretch{1.2}
\centering
\caption{\centering Ablation study of BEC module}
\label{table6}
\begin{tabular}{c|ccc}
\hline
            & \begin{tabular}[c]{@{}c@{}}DVSCLEAN\\ SNR\end{tabular} & \begin{tabular}[c]{@{}c@{}}DVSNOISE20\\ RPMD\end{tabular} & \begin{tabular}[c]{@{}c@{}}ED-KoGTL\\ SNR\end{tabular} \\ \hline
without BEC & 21.65                                                    & 35.49                                                       & 22.31                                                    \\
with BEC    & 25.73                                                    & 28.12                                                       & 26.97                                                    \\ \hline
\end{tabular}
\end{table}

\textbf{Effect on the subsequent task.} Event denoising is a kind of low-level task aiming to facilitate the following tasks, such as object classification and gesture recognition. The degree of improvement on the subsequent task is a significant metric to evaluate the denoising performance. To prove the effect of our \ourmethod{} on object classification, we compare classification accuracy between the original data and the denoised data using HATS on MNIST-DVS~\cite{berner2013240x180}, N-CARS~\cite{sironi2018hats} and CIFAR10-DVS~\cite{li2017cifar10}.
HATS is one of the SOTA object classification algorithms, which transforms events to histograms of averaged time surfaces and then is fed to a support vector machine for inference. MNIST-DVS and CIFAR10-DVS datasets are the DVS version of the popular frame-based dataset, MNIST~\cite{lecun1989handwritten} and CIFAR10~\cite{krizhevsky2009learning}. They are recorded by moving the images of monitors in front of a fixed camera. N-CARS is directly recorded by event-based camera in urban environments.

\begin{table}[t]
\setlength{\abovecaptionskip}{2pt}
\renewcommand\arraystretch{1.2}
\centering
\caption{\centering Classification accuracy comparison between the original data and the denoised data using HATS.}
\label{table5}
\begin{tabular}{c|ccc}
\hline
        & \text{MNIST-DVS} & \text{N-CARS} & \multicolumn{1}{l}{\text{CIFAR10-DVS}} \\ \hline
\text{Raw data}                                                                      & 98.4\%             & 90.2\%          & 52.4\%                                   \\
\text{Denoised data}                                                                 & 99.1\%             & 92.1\%          & 60.1\%                                   \\ \hline
\text{Added value}                                                                   & 0.7\%              & 1.9\%           & 7.7\%                                    \\ \hline
\text{\begin{tabular}[c]{@{}c@{}}Added value/(100\%\\ -original value)\end{tabular}} & 43.8\%             & 19.4\%          & 16.2\%                                   \\ \hline
\end{tabular}
\end{table}

The comparison of the classification accuracy rates before and after denoising can be seen in Table~\ref{table5}. The accuracy increases $0.7\%$, $1.9\%$ and $7.7 \%$, respectively. The improvement of N-CARS and CIFAR10-DVS is remarkable. The minor improvement on MNIST-DVS is due to the especially high original accuracy, which makes it hard to make progress. For these datasets with a high accuracy rate using HATS, we use the ratio of added value and (100$\%$ original value) to verify the validity. This criterion is able to reflect the relative increase in classification accuracy after denoising. The result of MNIST-DVS overcomes 40$\%$, which proves the effectiveness of our \ourmethod{}.

\section{Conclusion}
\label{5}
In this work, we give the theoretical analysis of separating real-world events from noisy event stream and establish Temporal Window (TW) module and Soft Spatial Feature Extraction module to process spatial and temporal information based on the analysis separately. Then, we propose a multi-scale window-based event denoising neural network, named \ourmethod{}, which aims to improve denoising efficiency. Hierarchical Spatial Feature Learning (HSFL) structure and Bone Events Check (BEC) unit are used to mitigate the passive impact on denoising accuracy brought by the increase in the number of processing events. The experimental performance of our \ourmethod{} shows better event denoising ability compared to other SOTA algorithms.

\section*{Acknowledgment}
This work was partially supported by NSFC under Grant 62022063 and the National Key R$\&$D Program of China under Grant 2018AAA0101400.

\newpage

\vfill

\end{document}